\documentclass[conf]{new-aiaa}
\usepackage[utf8]{inputenc}

\usepackage{graphicx}
\usepackage{amsmath}
\usepackage{algorithm}
\usepackage{algpseudocode}
\usepackage[version=4]{mhchem}
\usepackage{siunitx}
\usepackage{longtable,tabularx}
\usepackage{subcaption}

\usepackage{hyperref}

\algrenewcommand\algorithmicrequire{\textbf{Input:}}
\algrenewcommand\algorithmicensure{\textbf{Output:}}
\algnewcommand\Break{\textbf{break}} 

\title{Task Allocation and Motion Planning in Dynamic, Cluttered Environments via CBBA and Graphs of Convex Sets}

\author{Matthew D. Osburn\footnote{Research Assistant, Electrical and Computer Engineering, osburnm@student.byu.edu, AIAA Student Member.}, Cameron K. Peterson\footnote{Professor, Electrical and Computer Engineering, cammy.peterson@byu.edu, AIAA Senior Member.}, and John L. Salmon\footnote{Professor, Mechanical Engineering, johnsalmon@byu.edu, AIAA Member.}}
\affil{Brigham Young University, Provo, Utah, 84602}

\begin{document}

\maketitle

\begin{abstract}
Multi-agent task planning in cluttered, dynamic environments requires assigning tasks to agents while simultaneously determining safe, time-efficient trajectories through the environment. When tasks are dynamic, such as rendezvous objectives, allocation decisions depend not only on which agent is best suited for a task, but also on when and where that task can be reached.

This paper presents a solution to this problem, which combines Graphs of Convex Sets (GCS) for trajectory optimization with the Consensus-Based Bundle Algorithm (CBBA) for distributed task allocation. In our approach, GCS finds optimal trajectories through dynamic environments using a time-extended (3D+time) configuration space. At the same time, CBBA coordinates task assignments across agents, enabling informed decision-making in a moving environment. We then connect allocation and planning to allow the agents to avoid collisions in the 3D+time configuration space and provide accurate time estimates for task completion.  We demonstrate the effectiveness of our approach in simulated cluttered environments with static and dynamic tasks. 
\end{abstract}

\section{Introduction}

\lettrine{M}{ulti}-agent systems operating in complex environments must reason jointly about task assignment and motion feasibility~\cite{antonyshyn_multiple_2023}. In multi-task scenarios, agents must decide which tasks to execute while also generating trajectories that satisfy dynamic constraints~\cite{savla_traveling_2008} and avoid obstacles~\cite{kim_adversarial_2019}. These requirements create a coupled allocation--planning problem in which task utility depends not only on assignment decisions, but also on whether the assigned tasks can be physically realized.

Motion planning plays a critical role in multi-agent task allocation. In realistic settings such as dense obstacle fields, narrow passages, and time-varying environments, the cost and feasibility of executing a task must be accurately coupled with the task allocation scheme~\cite{hauser_integrating_nodate}. Ignoring these constraints during allocation can produce assignments that are infeasible or significantly suboptimal when executed~\cite{bidot_geometric_2017}. This motivates approaches that more tightly integrate motion planning into the decision-making process, enabling allocation strategies that are aware of geometric and temporal constraints~\cite{motes_multi-robot_2020}.

Graphs of Convex Sets (GCS) have emerged as a powerful framework for motion planning that bridges discrete and continuous optimization.  By decomposing the configuration space into convex regions and connecting them into a graph structure~\cite{akin_computing_2015}, optimal collision-free trajectories can be planned without requiring an initial guess from the end user~\cite{marcucci_shortest_2023}. This formulation uses mixed-integer convex programming, which quickly finds globally optimal trajectories through the provided graph~\cite{morozov_multi-query_2024, garg_planning_2024}.
The GCS framework has been successfully applied to a variety of motion planning problems, including collision-free path planning~\cite{marcucci_motion_2023}, high-dimensional motion planning~\cite{marcucci_graphs_2024}, and non-Euclidean planning domains~\cite{cohn_non-euclidean_2024}.

Recent work, including our own, has extended GCS to 2D+time spatiotemporal domains, named Space-Time Graphs of Convex Sets (ST-GCS)~\cite{tang_space-time_2025,osburn_systematic_2025}.
ST-GCS introduces time as an explicit dimension, enabling the generation of collision-free, time-parameterized trajectories in dynamic environments~\cite{tang_space-time_2025}. By covering the space-time domain with convex sets, ST-GCS formulations incorporate velocity bounds, timing constraints, and dynamic obstacles within a unified optimization framework~\cite{osburn_systematic_2025}. This formulation avoids the non-linear timing optimization step that was required in the prior GCS papers~\cite{marcucci_motion_2023, marcucci_shortest_2023, cohn_non-euclidean_2024,morozov_multi-query_2024, marcucci_motion_2022}.
This paper extends ST-GCS to incorporate an additional dimension, allowing for 3D+time trajectory planning.

The GCS literature has also briefly addressed multi-agent motion planning. In~\cite{garg_planning_2024}, the motion of two robotic arms was planned by combining and simplifying the configuration space of the systems and then using GCS to plan the combined motion.  In~\cite{tang_space-time_2025}, the authors used a priority planning queue to alternate between planning the motion of agents and updating the representation of the collision-free space. 
To our knowledge, GCS has not been used as the motion planner within complex, decentralized, task-allocation frameworks, such as the Consensus-Based Bundle Algorithm (CBBA).

CBBA is a widely used approach to multi-agent task allocation because it allows each agent to construct a locally ordered task bundle (the ordered subset of tasks that the agent intends to complete), exchange bids with its neighbors, and converge to a conflict-free assignment without centralized optimization~\cite{choi2009cbba}. Since these bundles are formed by selecting tasks according to their bid values, the resulting assignment is shaped by the cost or reward model used to compute those bids. In this work, we focus on settings where simplified bid models, such as geometric distance or nominal travel time, may not reflect the trajectory-level constraints that determine whether a task can be executed safely and efficiently.

Several works have already explored tighter couplings between CBBA and motion generation. CBBA has been combined with Rapidly-Exploring Random Tree (RRT) planning in~\cite{luders_information-rich_2011} to create collision-free dynamically feasible task allocations. CBBA can be modified to allow for real-time re-planning in dynamic environments and can be tailored for heterogeneous swarms of agents as in~\cite{ponda_decentralized_2010}. A more recent CBBA-based UAV mission-planning study extended the allocation loop to account for timing constraints, asynchronous reassignment, dynamic task insertion, and local replanning around the current execution path~\cite{chen_consensus-based_2022}. These results show that CBBA can absorb increasingly path- and time-dependent decision logic, but the motion layer is still often based on sampling-based or heuristic replanning, rather than one that can provide optimality guarantees.

In this paper, we use ST-GCS as the planning backend for CBBA in cluttered, dynamic multi-agent task-allocation scenarios.  Our first contribution is a 3D+time GCS formulation that explicitly plans over three spatial dimensions and time, extending existing space-time graph-of-convex-sets methods beyond lower-dimensional dynamic planning (2D+time) settings. Our second contribution is a motion-aware CBBA framework in which ST-GCS is used in the bidding process to compute task costs from convex trajectory optimization rather than from simplified distance or travel-time approximations.   This integration allows allocation decisions to account directly for dynamic obstacles, temporal feasibility, and trajectory-level constraints.
A third contribution is the extension of the task-allocation problem to include dynamic tasks, allowing task locations and availability to vary over time rather than assuming a fixed static location. In this paper, we interpret dynamic tasks as a rendezvous between agents and other vehicles that are not part of the planning cohort, but the same formulation applies to target interception, mobile sensing or inspection, and logistics handoff with moving assets.
Finally, we evaluate ST-GCS as a distributed motion-planning backend for collision-free motion planning in dynamic environments with accurate task-completion estimates.

The remainder of the paper is organized as follows. Section~\ref{sec:background} reviews the relevant background on CBBA and GCS. The proposed ST-GCS-CBBA framework is developed in Section~\ref{sec:methods}, including the integration of task allocation and motion planning. Section~\ref{sec:results} evaluates the approach through representative case studies. Finally, Section~\ref{sec:conclusion} provides a conclusion and discussion of limitations and future work.

\begin{figure*}[t]
\centering
\includegraphics[]{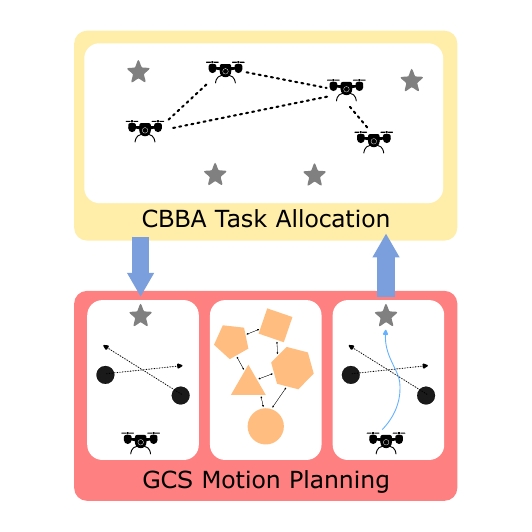}
\caption{Coupled CBBA and GCS for Task Allocation. Distributed agents in a graph are allocated tasks, shown as grey stars, by the CBBA algorithm. The Graphs of Convex Sets (GCS) framework is used as the motion planner within a dynamic environment to plan trajectories and provide accurate information for informed CBBA bids.}
\label{fig:cbba_gcs_overview}
\end{figure*}

\section{Background} \label{sec:background}

This section reviews the two key algorithmic components that underlie the proposed framework. We first summarize CBBA, which provides distributed task allocation over a finite task set, and then review GCS, which provides a motion-planning framework over a convex decomposition of the feasible domain. Figure~\ref{fig:cbba_gcs_overview} provides a high-level view of how these components interact in the overall planning pipeline. The objective is not to provide an exhaustive summary of both methods, but to establish the assumptions, terminology, and solution properties needed to understand the method developed in Section~\ref{sec:methods}.  For a more comprehensive overview of these two topics see~\cite{choi2009cbba} for CBBA and ~\cite{marcucci_shortest_2023} for GCS.

\subsection{Consensus-Based Bundle Algorithm}

\begin{figure}[t]
\centering
\includegraphics[width=\linewidth]{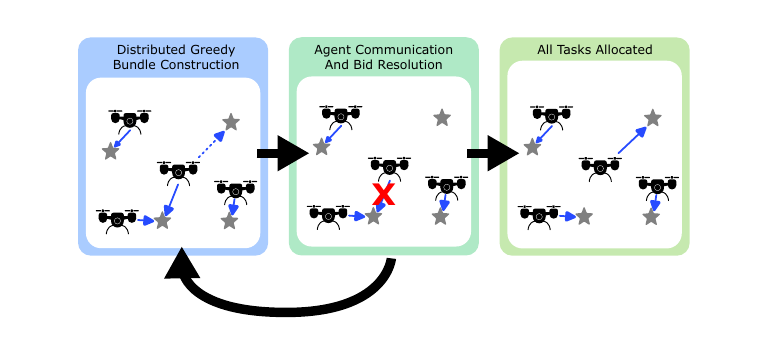}
\caption{In CBBA, agents alternate between individual greedy local bundle construction and consensus-based bid resolution with neighboring agents until all the tasks are allocated.}
\label{fig:cbba_overview}
\end{figure}

CBBA is a distributed auction-based method for multi-agent, multi-task allocation originally introduced in~\cite{choi2009cbba}. Each agent incrementally constructs an ordered bundle of tasks together with an associated execution path by repeatedly selecting the task insertion that yields the largest positive marginal score. That score may reflect reward, travel cost, timing penalties, or fuel use, which makes CBBA well-suited to path-dependent allocation problems in which task value depends on execution order rather than on independent one-step assignments. After local bundle construction, agents exchange bid and winner information with their neighbors and apply a consensus step to resolve conflicts. Repeated bundle building and consensus produce a conflict-free assignment without requiring a centralized auctioneer. Figure~\ref{fig:cbba_overview} depicts this process.

The first phase of CBBA constructs a candidate bundle of tasks for each agent using a greedy insertion rule. During this phase, each agent independently evaluates every task that is not already in its bundle and computes the marginal improvement obtained by inserting that task into every feasible position in its current execution path. The agent then bids on the task that produces the largest valid marginal gain, provided that this bid exceeds the agent's current estimate of the winning bid for that task. The selected task is appended to the agent's bundle, inserted into the best position in its execution path, and the corresponding winning bid and winning-agent estimates are updated. This process repeats until the agent reaches its task capacity or no remaining tasks can be won.  This process is outlined in Algorithm~\ref{alg:greedy_cbba}.

\begin{algorithm}
\caption{\(\mathrm{CBBA}\) Greedy Bundle Construction}
\begin{algorithmic}[1]
\Require \(\mathcal A=\{0,1,\dots,N_{\text{agents}}-1\},\ \mathcal T=\{0,1,\dots,M_{\text{tasks}}-1\},\ L_{\text{capacity}}\in\mathbb N\)
\Require \(\mathbf b_i,\mathbf p_i,\mathbf y_i,\mathbf z_i,\ \forall i\in\mathcal A\)
\Ensure \(\mathbf b_i,\mathbf p_i,\mathbf y_i,\mathbf z_i,\ \forall i\in\mathcal A\)

\For{\(i\in\mathcal A\)}
    \While{\(|\mathbf b_i|<L_{\text{capacity}}\)}
        \For{\(j\in\mathcal T\)}
            \State \(c_{ij}\gets -\infty\)
            \State \(n_{ij}^{*}\gets 0\)
            \If{\(j\not\in\mathbf b_i\)}
                \State \( \Delta_k \gets \emptyset \)

                \For{\(k\in\{0,1,\dots,|\mathbf p_i|\}\)}
                    \State \(\Delta_k
                    \gets \Delta_k \oplus_{\text{end}}
                    S_i(\mathbf p_i\oplus_k j)-S_i(\mathbf p_i)\)

                \EndFor
                \State \( c_{ij} \gets \text{max}_k \text{ } \Delta_{k} \)
                \State \( n_{ij}^{*} \gets \text{argmax}_k \text{ } \Delta_{k} \)
            \Else
                \State \(c_{ij}\gets 0\)
                \State \(n_{ij}^{*}\gets \emptyset\)
            \EndIf
        \EndFor

        \State \( h_{ij} \gets \mathbb{I}(c_{ij} > y_{ij}) \quad \forall j \in \mathcal{T} \)
        \State \( L_i \gets \text{argmax}_j \text{ }  c_{ij} \cdot h_{ij} \)
        \If{\( c_{i,L_i}h_{i,L_i}=0 \)}
            \State \textbf{break}
        \EndIf
        \State \(\mathbf b_i\gets \mathbf b_i\oplus_{\mathrm{end}} L_i\)
        \State \(\mathbf p_i\gets \mathbf p_i\oplus_{n_{iL_i}^{*}} L_i\)
        \State \(y_{i,L_i}\gets c_{i,L_i}\)
        \State \(z_{i,L_i}\gets i\)
    \EndWhile
\EndFor
\State \Return \(\mathbf b_i,\mathbf p_i,\mathbf y_i,\mathbf z_i,\ \forall i\in\mathcal A\)
\end{algorithmic}\label{alg:greedy_cbba}
\end{algorithm} 

The set of agents is denoted by
\(\mathcal A=\{0,1,\dots,N_{\text{agents}}-1\}\), where each element
\(i\in\mathcal A\) indexes one agent in the team. The set of tasks is denoted by
\(\mathcal T=\{0,1,\dots,M_{\text{tasks}}-1\}\), where each element
\(j\in\mathcal T\) indexes one candidate task. The scalar
\(L_{\text{capacity}}\) defines the maximum number of tasks that any one agent may place in its bundle.

For each agent \(i\), the vector \(\mathbf b_i\) denotes the agent's task bundle. The bundle is ordered according to the sequence in which tasks were added by the greedy construction process. Thus, if a task appears earlier in \(\mathbf b_i\), it was selected earlier by agent \(i\). The vector \(\mathbf p_i\) denotes the corresponding execution path. Unlike the bundle, the path is ordered according to the sequence in which agent \(i\) plans to execute the selected tasks. Therefore, \(\mathbf b_i\) records task-selection order, while \(\mathbf p_i\) records task-execution order.

The entry \(y_{ij}\) is agent \(i\)'s current estimate of the winning bid for task \(j\). This value represents the highest bid for task \(j\) known to agent \(i\) at the current point in the algorithm. The entry \(z_{ij}\) is agent \(i\)'s current estimate of the winning agent for task \(j\). Thus, \(z_{ij}=r\) means that, according to agent \(i\)'s local information, agent \(r\) is currently winning task \(j\).

The variable \(c_{ij}\) is the marginal score, or bid, that agent \(i\) computes for task \(j\). It measures the improvement in agent \(i\)'s path score if task \(j\) is inserted into the best available position in the current path \(\mathbf p_i\). The insertion index is denoted by \(k\), where
\(k\in\{0,1,\dots,|\mathbf p_i|\}\). Since the algorithm is zero-indexed, \(k=0\) corresponds to inserting the task at the beginning of the path, while \(k=|\mathbf p_i|\) corresponds to appending the task to the end of the path.

The operator \(\oplus_k\) denotes insertion at index \(k\). For example,
\(\mathbf p_i\oplus_k j\) is the path obtained by inserting task \(j\) into path \(\mathbf p_i\) at position \(k\). The operator \(\oplus_{\mathrm{end}}\) denotes appending an element to the end of a list. Thus,
\(\mathbf b_i\oplus_{\mathrm{end}}L_i\) is the updated bundle obtained by appending the selected task \(L_i\) to the end of \(\mathbf b_i\).

For each candidate task \(j\), the algorithm evaluates the score of the associated path \(S_i(\cdot)\) and the score improvement associated with each feasible insertion index \(k\). These improvements are stored in \(\Delta_k\), where
\(
\Delta_k = S_i(\mathbf p_i\oplus_k j)-S_i(\mathbf p_i).
\)
The best marginal score for task \(j\) is then
\(
c_{ij} = \max_k \Delta_k,
\)
and the corresponding best insertion index is
\(
n_{ij}^{*} = \arg\max_k \Delta_k
\). This process quantifies how much additional value agent \(i\) obtains by inserting task \(j\) into position \(k\) of its current path.
After all candidate tasks have been evaluated, the binary variable \(h_{ij}\) indicates whether task \(j\) is currently winnable by agent \(i\):
\(
h_{ij}=\mathbb I(c_{ij}>y_{ij}).
\)
If \(h_{ij}=1\), then agent \(i\)'s bid for task \(j\) exceeds its current estimate of the winning bid. If \(h_{ij}=0\), then agent \(i\) does not currently outbid the known winner for that task.

The selected task is denoted by \(L_i\), where
\(
L_i = \arg\max_j c_{ij}h_{ij}.
\)
This selects the task with the largest valid marginal score. If the best valid score satisfies
\(
c_{i,L_i}h_{i,L_i}=0,
\)
then no remaining task can be profitably added to the bundle, and the greedy construction process terminates for agent \(i\). Otherwise, \(L_i\) is appended to the bundle, inserted into the path at \(n_{iL_i}^{*}\), and the winning bid and winning-agent estimates are updated according to
\(
y_{i,L_i}\gets c_{i,L_i},\text{ }
z_{i,L_i}\gets i.
\)

CBBA is widely used in distributed coordination problems because it scales well with agent count and can accommodate heterogeneous agents, path-dependent utilities, and communication-limited execution. Its standard formulation assumes a finite task set, a strongly connected communication network over which bid and winner information eventually propagates, and a scoring rule that permits meaningful marginal evaluation of candidate task insertions. In practice, the quality of the final allocation depends strongly on the fidelity of the bidding model: if motion costs are oversimplified, the resulting assignment may remain conflict-free yet still be inefficient or infeasible once detailed planning constraints are imposed.

CBBA does not, in general, guarantee a globally optimal allocation. The bundle-building step is greedy, and agents reason locally during each consensus round. Its principal theoretical benefit is finite-time convergence to a consistent assignment under the standard communication and bidding assumptions. In the present work, CBBA is therefore used as a scalable distributed coordination layer whose performance depends on the accuracy of the ST-GCS planner.


\begin{figure}[t]
\centering
\includegraphics[width=\linewidth]{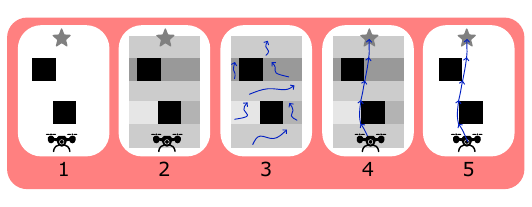}
\caption{The graphs of convex sets problem is solved by 1) establishing the start and end of the trajectory and determining the location of obstacles (shown in black), 2) segmenting collision free regions of free space into convex sets, 3) assigning small segments of the trajectory to each convex set (shown in blue), 4) solving the GCS problem for the optimal trajectory through the graph, 5) the trajectory is optimal with respect to the graph and is collision free. }
\label{fig:gcs_overview}
\end{figure}

\subsection{Graphs of Convex Sets}

GCS provides a trajectory optimization framework for solving certain nonlinear, multimodal planning problems to global optimality with respect to a specified graph~\cite{marcucci_shortest_2023}. An advantage that GCS has over other optimization and sampling methods is that the user does not have to specify an initial guess for the trajectory to find the global optimal solution.  The method represents the planning problem as a shortest-path problem over a graph whose vertices correspond to convex subsets of the collision-free configuration space. Although the resulting formulation is naturally mixed-integer, GCS has a tight convex relaxation that can be solved efficiently. 

In practice, the graph is constructed by covering the collision-free configuration space with convex regions, adding edges between adjacent or overlapping regions, and optimizing a trajectory that passes through a sequence of these sets. The Iterative Region Inflation by Semidefinite Programming (IRIS) family of algorithms, used in~\cite{akin_computing_2015, petersen_growing_2023,werner_faster_2024}, is commonly used to generate these convex regions in higher-dimensional configuration spaces. 

An abstract form of the GCS optimization problem is
\begin{subequations} \label{eq:gcs}
\begin{align}
 \min_{x} \quad & \sum_{v \in \mathcal{V}} f_{v}(\vec{x}_{v}) + \sum_{e \in \mathcal{E}} f_{e}(\vec{x}_{a}, \vec{x}_{b}) \\
 \text{s.t.} \quad \notag \\ 
 & (\mathcal{V},\mathcal{E}, \mathcal{C}) \subseteq \mathcal{G} \\
&\vec{x}_{v} \in \mathcal{X}_{v} & \forall v\in \mathcal{V}\\
&\vec{x}_{a} \in \mathcal{X}_{e} & \forall e\in\mathcal{E}\\
&\vec{x}_{b} \in \mathcal{X}_{e} & \forall e\in\mathcal{E},
\end{align}
\end{subequations}
where $x$ represents all the continuous variables defining the trajectory curve segments, often Bézier control points. The variables $\vec{x}_{v}$ are the design variables associated with the vertex $v$. The vertex cost function $f_{v}(\vec{x}_{v})$ represents the cost of the continuous variables within a vertex. The edge cost function $f_{e}(\vec{x}_{a}, \vec{x}_{b})$ represents the cost of traversing an edge from $v_a$ to $v_b$, and couples the continuous variables of the connecting vertices. 

The vertices $\mathcal{V}$, edges $\mathcal{E}$, and associated convex sets $\mathcal{C}$ that form the optimal trajectory are a subset of the graph $\mathcal{G}$ that minimizes the overall cost. The sets $\mathcal{X}_v$ and $\mathcal{X}_e$ define the vertex and edge constraints that the design variables must satisfy. Often, the formulation is rewritten with binary variables and transformed to take the form of a mixed-integer convex program (MICP) that can be repeatedly solved to find the global solution to the original problem. 

Figure~\ref{fig:gcs_overview} illustrates the GCS framework, beginning with a planning problem defined by a start state, a goal state, and obstacles in the environment. The collision-free portion of the workspace is then decomposed into a collection of convex sets, which form the vertices of a graph. Candidate trajectory segments are assigned to these convex regions, allowing the trajectory search to be posed as an optimization over both the sequence of convex sets and the continuous trajectory variables within them. Solving the resulting GCS problem produces a collision-free trajectory that is optimal with respect to the constructed graph.

\begin{figure}[t]
\centering
\includegraphics[width=0.35\linewidth]{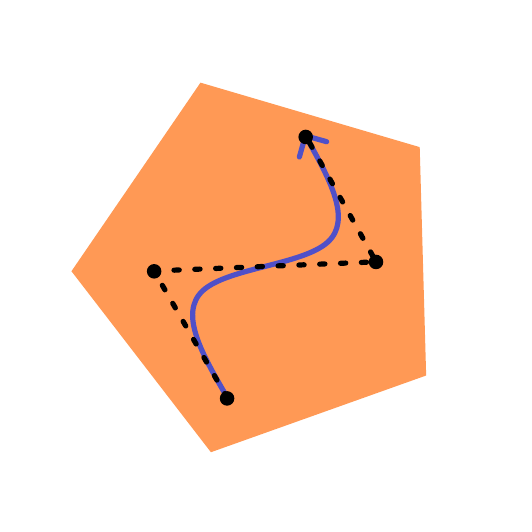}
\caption{The B\`ezier curve (blue) is entirely contained in the convex set (orange) if its connecting control points (black) are contained within the convex set.}
\label{fig:bezier_curve}
\end{figure}

In this paper, the trajectory segment within each convex set is parameterized using B\'ezier curves. A B\`ezier curve is defined as
\begin{align}
    \vec{B}(s) &= \sum_{k=0}^{n} \binom{n}{k}s^k(1-s)^{n-k}\vec{x}_k, \quad 0\leq s \leq 1,
\end{align}
where $s$ is a parameter that varies between $0$ and $1$, $n$ is the polynomial order of curve, and $x_k$ is the $k$th control point defining the curve.
This parameterization choice is useful because constraining all control points to lie inside a convex set ensures that the entire B\'ezier curve segment remains inside that set. Thus, collision avoidance can be enforced through convex containment constraints on the control points rather than through dense sampling along the trajectory. Figure~\ref{fig:bezier_curve} depicts this property. Adjacent curve segments are connected through edge constraints that enforce continuity between neighboring convex regions in the graph. A downside of this parameterization is the increased difficulty in enforcing feasibility for highly nonlinear vehicle dynamics.

When solved using branch and bound, or if the convex relaxation returns a complete solution, the resulting trajectory is globally optimal over the family of trajectories that can be represented with the pairing of B\`ezier curves and convex sets. In practice, using rounding procedures and sub-optimality metrics can substantially reduce computational cost. These approaches can yield strong candidate solutions close to or equal to the global optimum because of the tightness of the convex relaxation, but global optimality is no longer guaranteed.

The effectiveness of GCS depends on the quality of the convex decomposition and graph construction. The method assumes that feasible motion can be adequately represented by convex regions, and it inherits any flaws present in that representation. If important areas in the collision-free space are not covered within the graph, the trajectory will be optimal with respect to the provided graph, but not with respect to the actual collision-free space. This is an additional source of error that must be analyzed when evaluating the performance of GCS motion planning. All of these modeling considerations are especially important in cluttered environments, where narrow passages and obstacle geometry strongly influence the creation of the convex sets.

\begin{figure}[t]
\centering
\includegraphics[width=0.40\linewidth]{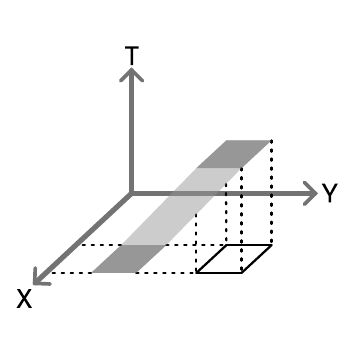}
\caption{A dynamic obstacle moving in the $y$ direction in $\mathbb{R}^2$ becomes a static parallel-piped in the time 2D+time configuration space.}
\label{fig:stgcs}
\end{figure}

For dynamic environments, the same idea can be extended into ST-GCS~\cite{tang_space-time_2025}. In ST-GCS, time is treated as an explicit planning dimension, so each convex region occupies a subset of the 2D+time domain rather than the spatial domain alone. Dynamic obstacles, timing constraints, and velocity limits can be encoded directly in the space-time configuration space.  An example of a dynamic obstacle in the 2D+time configuration space is depicted in Figure~\ref{fig:stgcs}.  It shows that a rectangular obstacle in 2D space becomes a parallelepiped in 2D+time dimensions.  ST-GCS ensures that collisions of dynamic obstacles are avoided.  The resulting solution of ST-GCS is both geometrically feasible and time-parameterized, which makes it well-suited to the coupled allocation-and-planning setting considered in this paper.

Existing CBBA extensions incorporate path-dependent or timing-aware bids, often through sampling-based planners, heuristic replanning, or simplified trajectory models, as in~\cite{chen_consensus-based_2022}. In parallel, GCS and ST-GCS methods provide strong trajectory optimization tools for planning through constrained environments, but they have not been tightly integrated with decentralized bundle-based task allocation in which each bid depends on solving a motion-planning problem. This paper addresses that interface by combining task allocation and trajectory optimization so that agents can form path-aware bids, navigate safely around obstacles, and reach both static and dynamic task objectives.

\section{Methods} \label{sec:methods}

The preceding sections motivate the need for a bid model that reflects trajectory-level feasibility rather than only geometric proximity or nominal travel time. In this section, we develop a CBBA+ST-GCS framework in which each agent evaluates candidate task insertions by solving space-time GCS motion-planning problems. This allows the CBBA bidding process to account directly for obstacle geometry, dynamic constraints, task timing, and trajectory cost during distributed bundle construction. We first describe how ST-GCS is embedded into the CBBA greedy bidding step, then present the 3D+time planning representation and the assumptions required to couple distributed task allocation with space-time motion planning.

\subsection{ST-GCS-Informed CBBA Greedy Bundle Construction}

The key coupling between CBBA and ST-GCS occurs in the marginal bid computation. In standard CBBA implementations, an agent often evaluates a candidate task insertion using a simplified surrogate cost, such as Euclidean distance, nominal travel time, or a hand-designed heuristic~\cite{choi2009cbba}. These surrogate models are computationally convenient, but they may not reflect whether the resulting task sequence is feasible in a cluttered, time-varying environment. In contrast, the proposed approach evaluates candidate insertions using the cost of a trajectory optimization problem solved in the same space-time domain in which the agent will execute its motion.

For each agent $i$, let $\mathbf{b}_i$ denote the ordered task bundle and let $\mathbf{p}_i$ denote the corresponding ordered execution path. The bundle records the order in which tasks are added during greedy construction, while the path records the order in which the agent plans to execute those tasks. During greedy bundle construction, agent $i$ considers each candidate task $j$ not already in its bundle and each possible insertion index $k$, producing a candidate path
\begin{equation}
    \mathbf{p}_i \oplus_k j .
\end{equation}
Rather than assigning a bid based only on the geometric distance added by this insertion, the agent evaluates the resulting execution sequence using ST-GCS. Each candidate path is decomposed into a sequence of point-to-point space-time planning problems between consecutive tasks or states. The resulting trajectory costs are then combined to evaluate the full path. Thus, each bid reflects not only task proximity, but also obstacle geometry, dynamic obstacles, timing constraints, velocity limits, and the feasibility of the resulting trajectory.

Because ST-GCS is formulated as a cost-minimization problem, we define the CBBA path score as a reward-minus-cost quantity. Let
\begin{equation} \label{eq:gcost}
    G_i(p_a,p_b)
\end{equation}
denote the optimal ST-GCS trajectory cost for agent $i$ to move from task or state $p_a$ to task or state $p_b$. The computation of the trajectory cost is described in detail in Section~\ref{sec:3d-gcs}.  The score for a candidate path is defined as
\begin{equation}
    S_i(\mathbf{p}_i)
    =
    -
    \sum_{\ell=0}^{|\mathbf{p}_i|-1}
    G_i(p_{\ell},p_{\ell+1}),
    \label{eq:path_score_reward_minus_cost}
\end{equation}
where $p_\ell$ represents the $\ell$th task in the candidate execution path.  The negative indicates that the minimum cost trajectory will be chosen. For cost maximization instead of minimization, the negative sign should be omitted in front of the sum. If any ST-GCS segment is infeasible, the full path is assigned
\begin{equation}
    S_i(\mathbf{p}_i) = -\infty ,
\end{equation}
so that infeasible insertions cannot win a bid.  Should all tasks and bundles be infeasible, then the agent is not assigned any tasks.

The marginal score associated with inserting task $j$ into path $\mathbf{p}_i$ at index $k$ is
\begin{equation}
    \Delta_{ijk}
    =
    S_i(\mathbf{p}_i \oplus_k j) - S_i(\mathbf{p}_i).
    \label{eq:marginal_score}
\end{equation}
The best bid that agent $i$ can place on task $j$ is then
\begin{equation}
    c_{ij}
    =
    \max_k \Delta_{ijk},
    \label{eq:best_bid}
\end{equation}
with corresponding insertion index
\begin{equation}
    n_{ij}^{*}
    =
    \arg\max_k \Delta_{ijk}.
    \label{eq:best_insertion_index}
\end{equation}
This preserves the standard CBBA greedy insertion structure, but replaces the local bid model with a trajectory-level optimization model. As a result, the task selected by CBBA is the task that provides the largest feasible improvement to the agent's space-time execution plan.

After all candidate tasks have been evaluated, agent $i$ determines whether its bid for task $j$ exceeds its current estimate of the winning bid for that task. Let $y_{ij}$ denote agent $i$'s current estimate of the winning bid for task $j$, and let $z_{ij}$ denote its current estimate of the winning agent. The binary variable
\begin{equation}
    h_{ij}
    =
    \mathbb{I}(c_{ij} > y_{ij})
\end{equation}
indicates whether task $j$ is currently winnable by agent $i$. The selected task is
\begin{equation}
    L_i
    =
    \arg\max_j c_{ij} h_{ij}.
\end{equation}
 Task $L_i$ is appended to the bundle and inserted into the execution path at $n_{iL_i}^{*}$, and the local winning-bid and winning-agent estimates are updated as
\begin{equation}
    y_{iL_i} \leftarrow c_{iL_i},
    \qquad
    z_{iL_i} \leftarrow i .
\end{equation}

This formulation is the main interface between distributed task allocation and motion planning in the proposed framework. CBBA remains responsible for decentralized task selection, conflict resolution, and consensus among agents, while ST-GCS supplies the feasibility-aware path scores that determine the bids. Each agent can therefore make local allocation decisions using motion plans that are consistent with the final trajectory-generation constraints. This is especially important in dynamic environments, where two task insertions with similar geometric distances may have very different space-time feasibility because of moving obstacles or task timing.

 In implementation, the ST-GCS solution associated with a candidate insertion may be performed for the full candidate path or only for the path segments affected by the insertion. For example, inserting a task in the middle of a bundle affects all tasks following, while appending a task changes only the final connection. Previously computed segment costs can therefore be cached and reused during greedy bundle construction. This preserves the CBBA structure while reducing the number of repeated ST-GCS solves required to evaluate candidate bids.

When agents exchange information during the consensus phase, they exchange the ST-GCS-informed bid values $c_{ij}$ and the corresponding winning-agent estimates $z_{ij}$. Once conflicts are resolved, the winning task sequence defines the agent's committed path.

\subsection{3D+Time ST-GCS Representation}\label{sec:3d-gcs}

The proposed CBBA bidding mechanism requires a motion planner that can evaluate not only geometric path length, but also time feasibility, dynamic obstacles, and task timing. To provide this capability, we extend the ST-GCS representation to a full 3D+time planning space. This extension allows each candidate CBBA task sequence to be evaluated using the same space-time constraints that will govern the executed trajectory, rather than using a simplified spatial approximation.

Each trajectory point is represented by a four-dimensional state,
\begin{equation}
    x_k =
    \begin{bmatrix}
        s_k \\
        \tau_k
    \end{bmatrix}
    \in \mathbb{R}^4,
    \label{eq:space_time_state}
\end{equation}
where $s_k \in \mathbb{R}^3$ is the spatial position of the multirotor and $\tau_k \in \mathbb{R}$ is the corresponding time coordinate. In this representation, time is not treated as a post-processing variable assigned after a spatial path is found. Instead, time is optimized directly as part of the trajectory. This allows moving obstacles, moving tasks, causality, and velocity limits to be represented within a single convex-set planning framework.

To enforce physically meaningful motion, consecutive trajectory control points are constrained by velocity and causality conditions. For control points $x_{v,i}$ and $x_{v,i+1}$ within a convex region, the maximum-speed constraint is
\begin{equation}
    \left\|x_{v,i+1}-x_{v,i}\right\|_{s,2}
    \leq
    v_{\max}
    \left|x_{v,i+1}-x_{v,i}\right|_{\tau},
    \label{eq:velocity_constraint_st}
\end{equation}
and the causality constraint is
\begin{equation}
    \left|x_{v,i+1}-x_{v,i}\right|_{\tau}
    \geq
    \lambda_{\epsilon}.
    \label{eq:causality_constraint_st}
\end{equation}
Here, $\|\cdot\|_{s,2}$ denotes the Euclidean norm over the spatial components and $|\cdot|_{\tau}$ denotes the signed magnitude of the temporal component. The constants $v_{\text{max}}$ and $\lambda_{\epsilon}$ represent the maximum velocity magnitude and the minimum time spacing between control points, respectively. The first constraint ensures that spatial displacement is consistent with the elapsed time between consecutive trajectory points. The second prevents degenerate solutions in which trajectory points collapse to the same time coordinate or move backward in time. Together, these constraints form a Lorentz-cone-type feasibility condition in 3D+time~\cite{boyd_convex_2023}.

For a single graph vertex, the trajectory cost combines distance, elapsed time, and smoothness:
\begin{equation}
    f_v(\mathbf{x}_v)
    =
    w_d G_{dist}(\mathbf{x}_v)
    +
    w_t G_{time}(\mathbf{x}_v)
    +
    w_s G_{smooth}(\mathbf{x}_v),
    \label{eq:vertex_cost}
\end{equation}
where $w_d$, $w_t$, and $w_s$ are nonnegative weights. The distance and time terms are
\begin{equation}
    G_{dist}(\mathbf{x}_v)
    =
    \sum_{i=0}^{n-1}
    \left\|
    \mathbf{x}_{v,i+1}-\mathbf{x}_{v,i}
    \right\|_{s,2},
    \label{eq:distance_cost}
\end{equation}
and
\begin{equation}
    G_{time}(\mathbf{x}_v)
    =
    \left|
    \mathbf{x}_{v,n}-\mathbf{x}_{v,0}
    \right|_{\tau}.
    \label{eq:time_cost}
\end{equation}
The smoothness term penalizes large changes in the spatial direction of the trajectory and discourages aggressive maneuvers that would be difficult for the multirotor to track. One suitable convex surrogate is
\begin{equation}
    G_{smooth}(\mathbf{x}_v)
    =
    \sum_{i=0}^{n-2}
    \left\|
    \mathbf{x}_{v,i+2}
    -
    2\mathbf{x}_{v,i+1}
    +
    \mathbf{x}_{v,i}
    \right\|_{s,2}.
    \label{eq:smoothness_cost}
\end{equation}
In this work, this term is used as a smoothness penalty rather than as a complete dynamic model. The ST-GCS planner enforces geometric collision avoidance, timing feasibility, and velocity bounds, while a lower-level tracking controller is assumed to execute the resulting sufficiently smooth reference trajectory.  The sum of all vertex costs $f_v$ over the graph at the end of optimization is the cost function $G(p_a,p_b)$ in Equation~\ref{eq:gcost}.

\begin{figure}[t]
\centering
\includegraphics[width=0.40\linewidth]{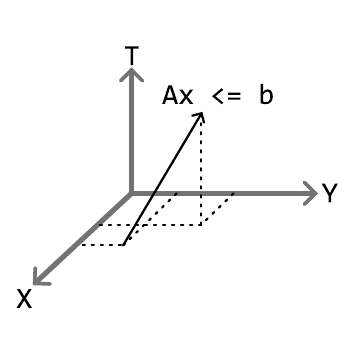}
\caption{The moving target (black arrow) can be formed by the $A$ and $b$ matrices.  This depiction is in $\mathbb{R}^3$, but is representative of constant velocity targets in higher dimensions.} 
\label{fig:stgcs_target}
\end{figure}

One major contribution of this paper is the introduction of dynamic tasks that can be naturally described in the ST-GCS framework. For static tasks, the terminal spatial region is fixed, and the arrival time is optimized within the planning horizon. For dynamic tasks, the terminal region changes with time. Both static and dynamic tasks can be generally represented by the constraint
\begin{equation}
Ax_f\leq b
\end{equation}
where
\begin{equation}
A = \begin{bmatrix}
    1&0&0&v_x \\
    -1&0&0&-v_x \\
    0&1&0&v_y \\
    0&-1&0&-v_y \\
    0&0&1&v_z \\
    0&0&-1&-v_z \\
\end{bmatrix},
b=\begin{bmatrix}
    x_{final}\\
    -x_{final}\\
    y_{final}\\
    -y_{final}\\
    z_{final}\\
    -z_{final}\\
\end{bmatrix},
\end{equation}
$x_f$ is the final control point of the ST-GCS trajectory, and $v_x, v_y, v_z$ are the 3D constant velocities of the task. If the task is static, then the velocity values will all be zero.  Figure~\ref{fig:stgcs_target} depicts the moving task defined by the $A$ and $b$ matrices.

The contribution of this formulation is not merely adding one more coordinate to the spatial planner. By treating time as an explicit optimization variable, the planner can determine both where and when the multirotor should move through the environment. This is essential for the proposed task-allocation framework because CBBA bids depend on the feasibility and cost of executing an ordered task sequence. A task that appears close in Euclidean space may be expensive or infeasible in space-time because of moving obstacles, moving task locations, or conflicts with other agents. The 3D+time ST-GCS representation allows these effects to be reflected directly in the bid values used during distributed allocation.

\section{Results} \label{sec:results}

This section evaluates the proposed CBBA+ST-GCS framework on two representative planning scenarios of increasing complexity. The first experiment isolates the ST-GCS planning component for a single multirotor assigned to three tasks. The second experiment evaluates the full multi-agent allocation and planning pipeline in cluttered dynamic environments with three agents and six tasks. Across all experiments, trajectories are generated in the 3D+time configuration space, and task bids are computed using the GCS-derived path score described in Section~III.

All experiments were run on a desktop computer with a 12th-generation Intel i7 CPU with 16GB of RAM. The implementation was written in PYTHON, with optimization problems modeled using CVXPY and solved using the open-source solver CLARABELL. All timing results report wall-clock computation time on the aforementioned machine.

\subsection{Experiment 1: Single multirotor with Three Tasks}

The first experiment considers a single multirotor assigned to complete three tasks in a 3D environment. Since only one agent is present, no inter-agent consensus or conflict resolution is required; instead, this case isolates the effect of the ST-GCS planner on greedy bundle construction. The multirotor begins at a fixed location and determines the task ordering that minimizes the total travel time while satisfying velocity and obstacle-avoidance constraints.

\begin{table}[b]
\centering

\begin{minipage}[t]{0.48\linewidth}
\centering
\caption{\label{tab:single_agent_setup} Experiment 1 setup}
\begin{tabular}{lc}
\hline
Parameter & Value \\\hline
Agents & 1 multirotor \\
Tasks & 3 \\
Maximum tasks per agent & 3 \\
Speed limit & 12 m/s \\
Planning horizon & 60 s \\
Static obstacles & 3 \\
Dynamic obstacles & 7 \\\hline
\end{tabular}
\end{minipage}
\hfill
\begin{minipage}[t]{0.48\linewidth}
\centering
\caption{\label{tab:single_agent_results} Experiment 1 results}
\begin{tabular}{lc}
\hline
Metric & Value \\\hline
Selected task order & East, West, North \\
Completed tasks & 3/3 \\
Total mission duration & 27.49s \\
\hline
\end{tabular}
\end{minipage}

\end{table}

\begin{figure}[t]
\centering
\begin{subfigure}[t]{0.49\linewidth}
    \centering
    \includegraphics[width=\linewidth]{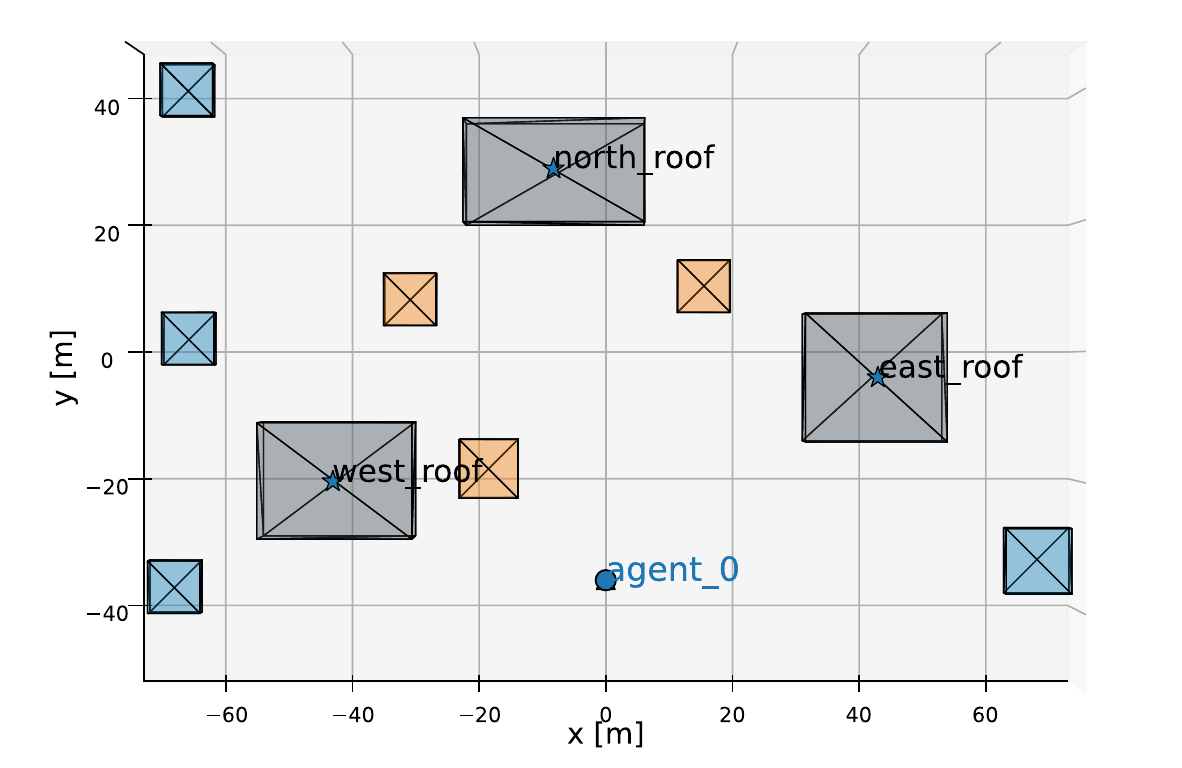}
    \label{fig:single_agent_environment_a}
\end{subfigure}
\hfill
\begin{subfigure}[t]{0.49\linewidth}
    \centering
    \includegraphics[width=\linewidth]{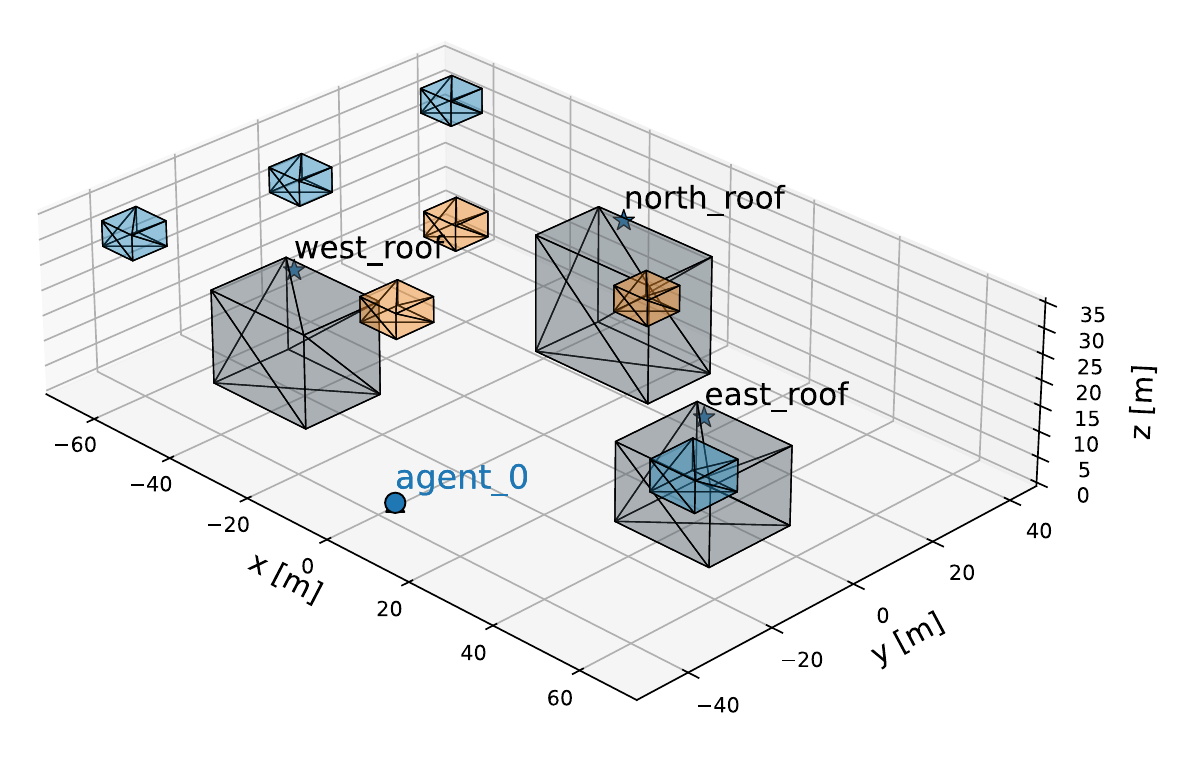}
    \label{fig:single_agent_environment_b}
\end{subfigure}
\caption{The agent must visit the three tasks shown as stars on the three buildings shown in grey. Slow dynamic obstacles, shown as orange boxes, and fast dynamic obstacles, shown as blue boxes, clutter the airspace.}
\label{fig:single_agent_environment}
\end{figure}

\begin{figure}[t]
\centering
\begin{subfigure}[t]{0.49\linewidth}
    \centering
    \includegraphics[width=\linewidth]{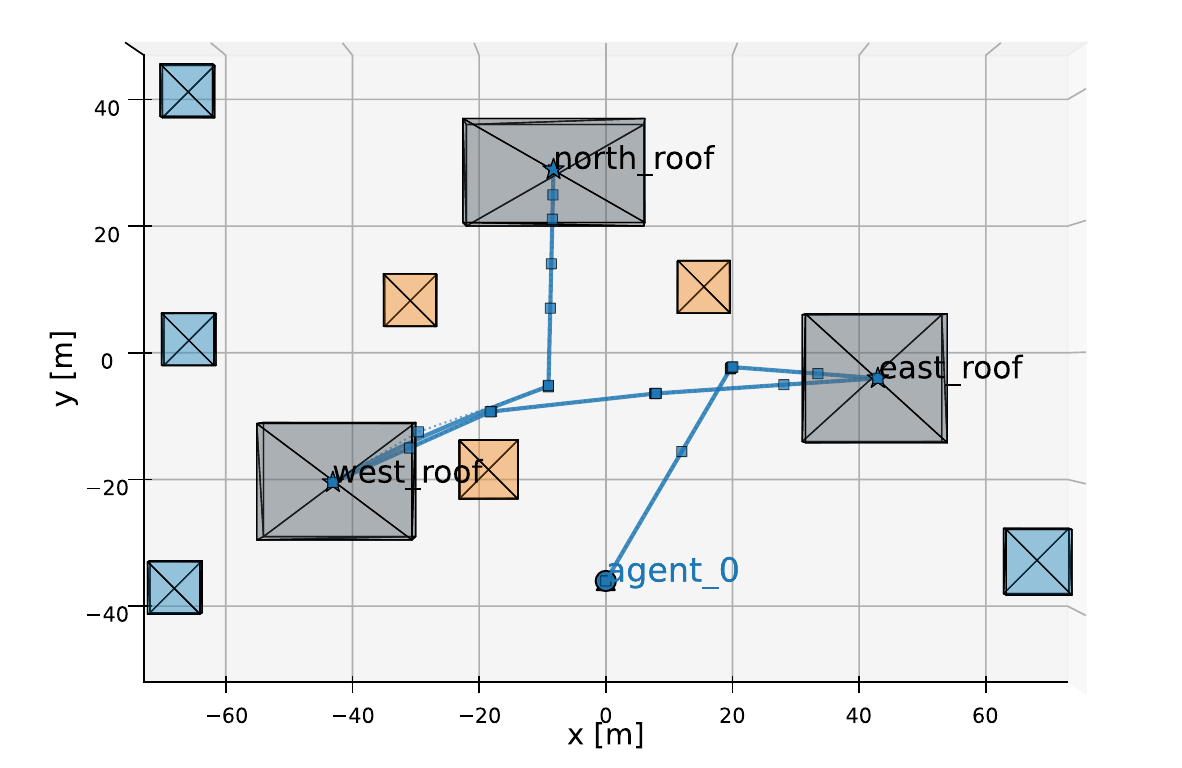}
    \label{fig:single_agent_environment_a}
\end{subfigure}
\hfill
\begin{subfigure}[t]{0.49\linewidth}
    \centering
    \includegraphics[width=\linewidth]{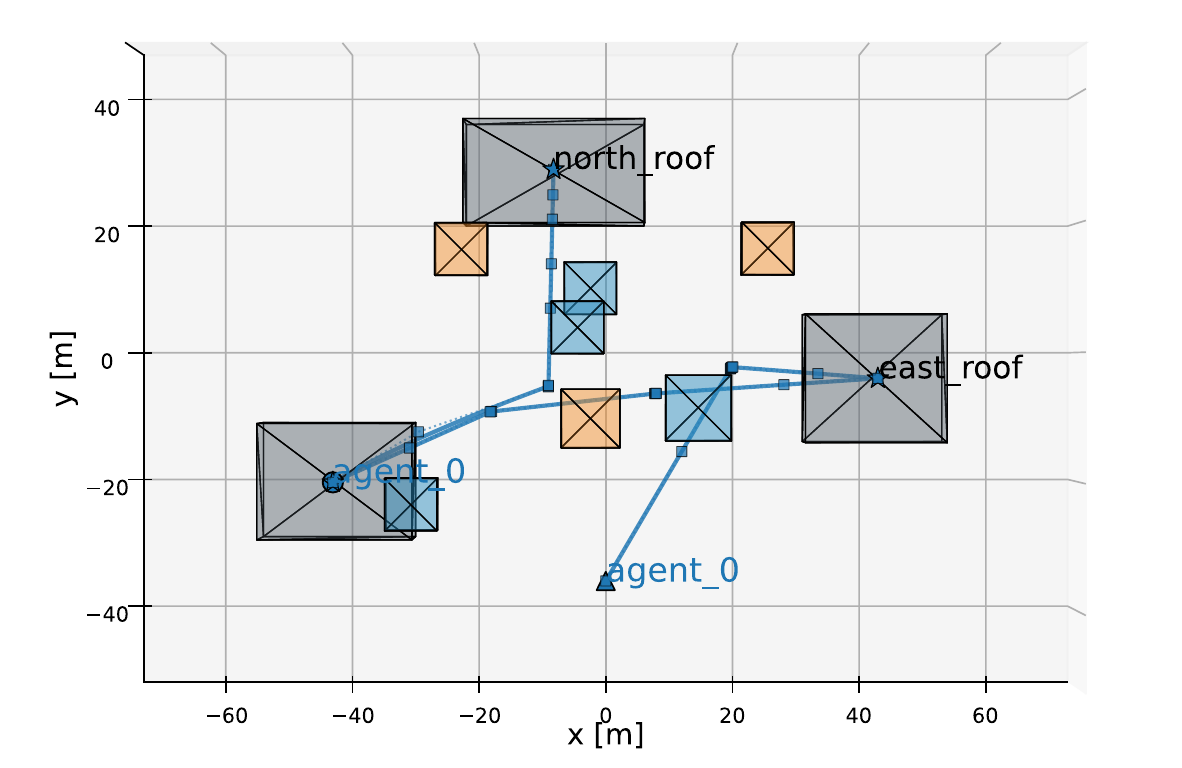}
    \label{fig:single_agent_environment_b}
\end{subfigure}
\caption{The CBBA+GCS successfully visits the three tasks while remaining collision-free (left). The dynamic obstacles make the airspace between tasks cluttered (right).}
\label{fig:single_agent_environment_result}
\end{figure}

Figure~\ref{fig:single_agent_environment} depicts the first scenario.
The environment contains static obstacles that represent buildings that the multirotor will visit. The environment contains sets of dynamic obstacles that either move quickly or slowly through the airspace, making traversal between tasks more difficult. For each candidate insertion considered during greedy bundle construction, the planner solves a sequence of GCS problems between consecutive tasks in the proposed path. The resulting travel time is used to evaluate the marginal score of adding the candidate task.  Table~\ref{tab:single_agent_setup} reports the parameters used for the experiment.

Figure~\ref{fig:single_agent_environment_result} shows the resulting collision-free trajectory created by the CBBA+GCS pipeline. 
Table~\ref{tab:single_agent_results} reports the success of this experiment along with the task assignment bundle that resulted from the CBBA algorithm and the total mission time. In this experiment, each task was successfully reached while each agent remained collision-free.

\subsection{Experiment 2: Static and Dynamic Tasks}

\begin{figure}[t]
\centering
\begin{subfigure}[t]{0.49\linewidth}
    \centering
    \includegraphics[width=\linewidth]{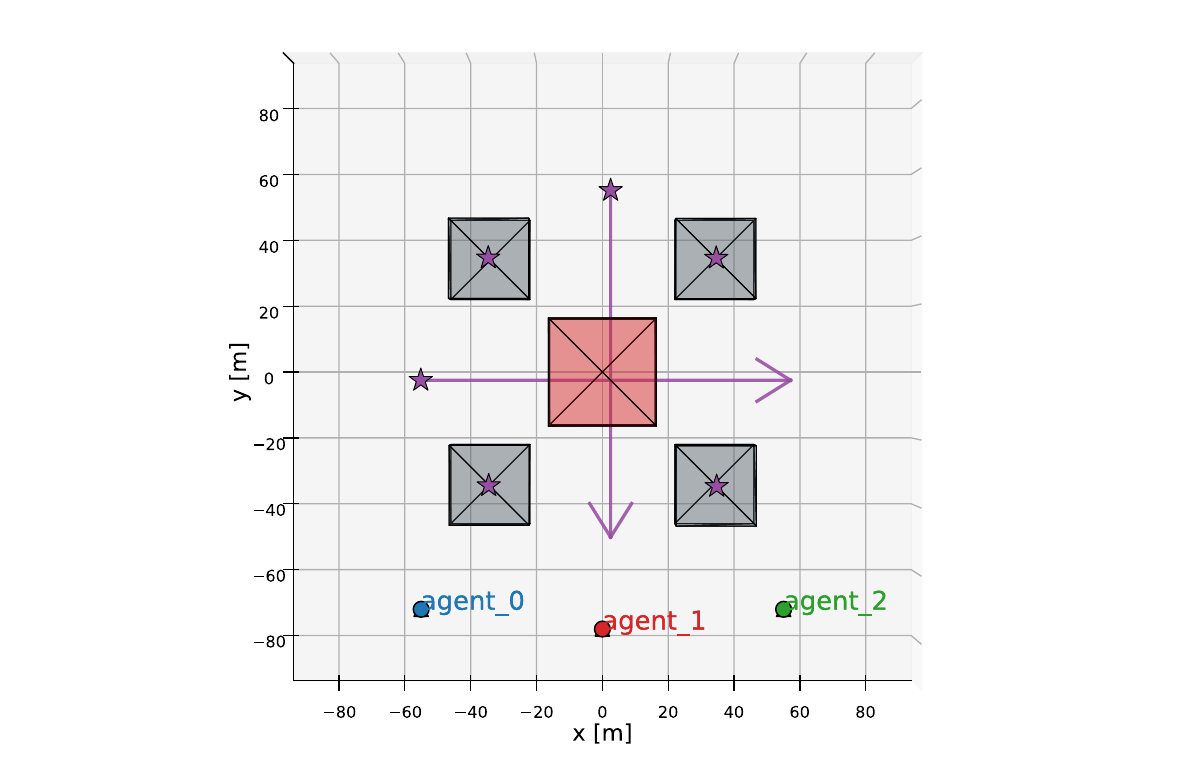}
    \label{fig:multi_agent_environment_a}
\end{subfigure}
\hfill
\begin{subfigure}[t]{0.49\linewidth}
    \centering
    \includegraphics[width=\linewidth]{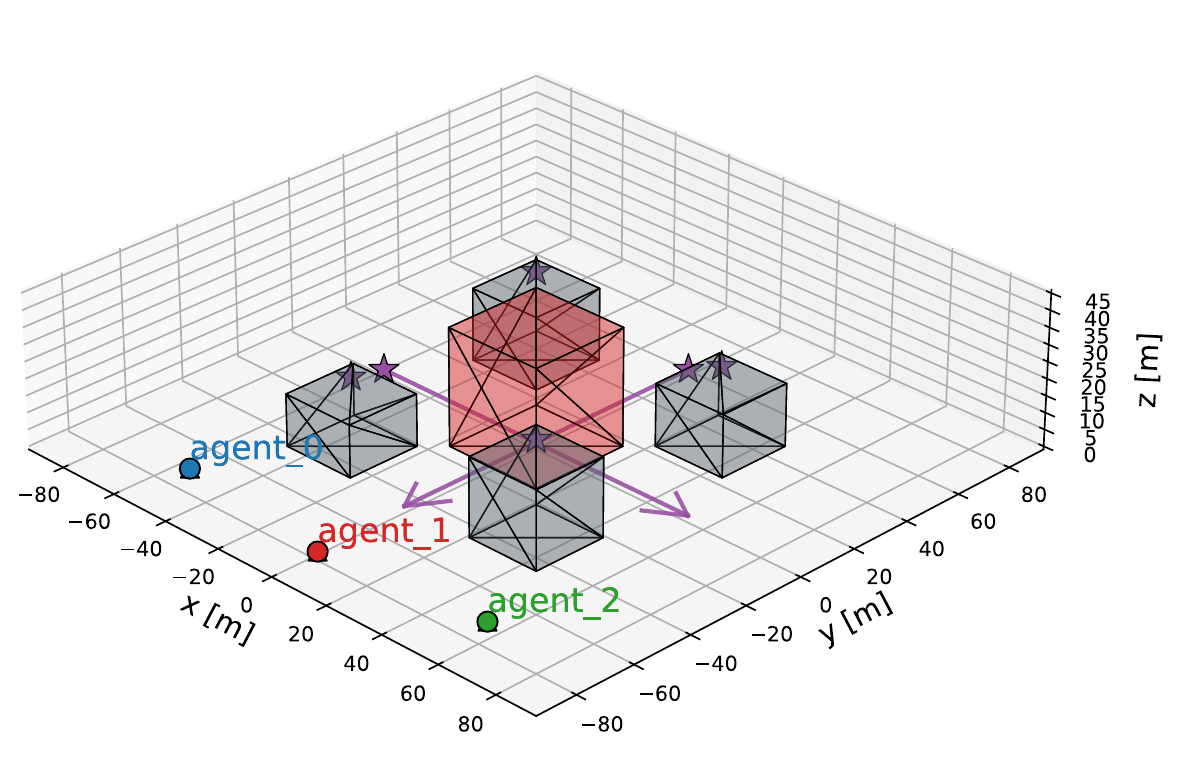}
    \label{fig:multi_agent_environment_b}
\end{subfigure}
\caption{The agents must visit the six tasks shown as purple stars at the intersection. Moving tasks are indicated by arrows.  There is a no-fly zone designated in red.}
\label{fig:multi_agent_environment}
\end{figure}

\begin{figure}[t!]
\centering
\begin{subfigure}[t]{0.49\linewidth}
    \centering
    \includegraphics[width=\linewidth]{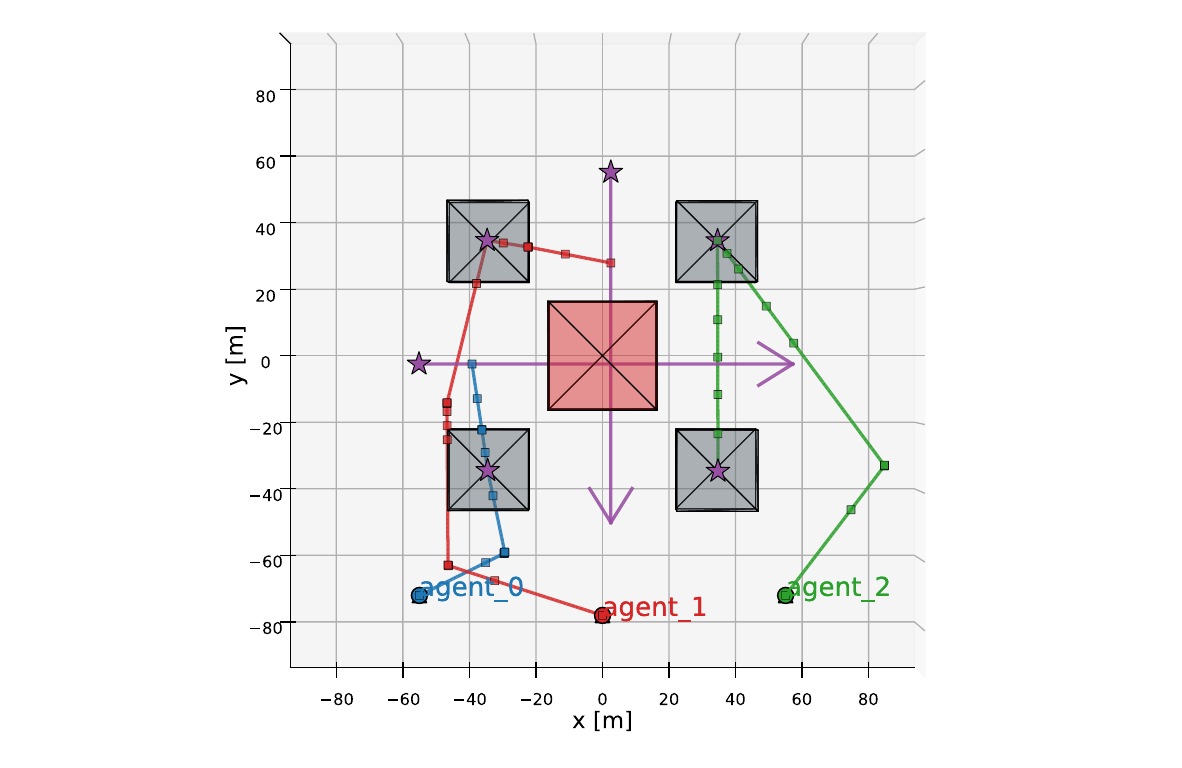}
    \label{fig:multi_agent_environment_result_a}
\end{subfigure}
\hfill
\begin{subfigure}[t]{0.49\linewidth}
    \centering
    \includegraphics[width=\linewidth]{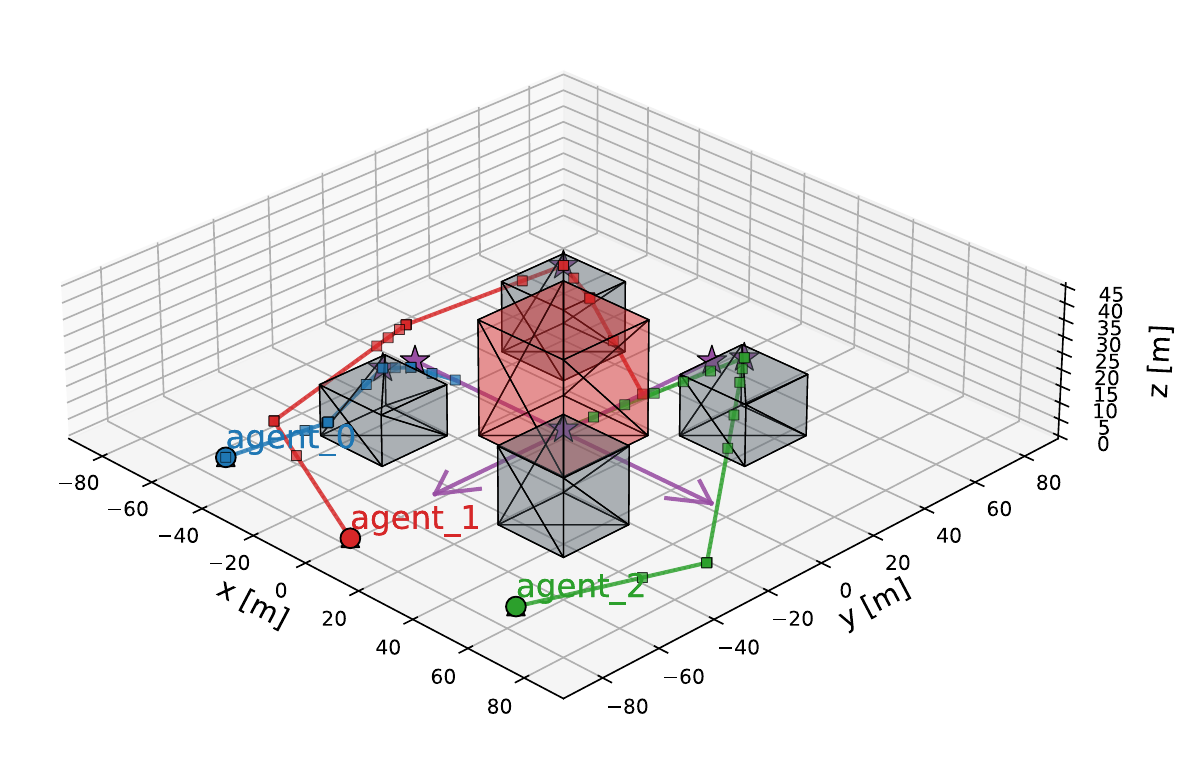}
    \label{fig:multi_agent_environment_result_b}
\end{subfigure}
\caption{The agents successfully distributed tasks using CBBA+GCS, including the time-varying tasks moving through the intersection, as well as avoiding the no-fly zone.}
\label{fig:multi_agent_environment_result}
\end{figure}

The second experiment evaluates the integrated CBBA+ST-GCS pipeline with three multirotors and six tasks.  Two of these tasks are moving in time through an intersection. Each multirotor is allowed to accept up to two tasks. Unlike the single-agent case, the allocation now depends on distributed bidding and consensus. Each agent computes GCS-informed bids for candidate task insertions, exchanges winner and bid information, and updates its bundle until a conflict-free assignment is reached.
\begin{table}[t]
\centering

\begin{minipage}[t]{0.48\linewidth}
\centering
\caption{\label{tab:multi_agent_setup} Experiment 2 setup}
\begin{tabular}{lc}
\hline
Parameter & Value \\\hline
Agents & 3 multirotors \\
Static tasks & 4 \\
Dynamic tasks & 2 \\
Maximum tasks per agent & 2 \\
Speed limit & 12 m/s \\
Planning horizon & 60 s \\
Static obstacles & 5 \\
Dynamic obstacles & 0 \\\hline
\end{tabular}
\end{minipage}
\hfill
\begin{minipage}[t]{0.48\linewidth}
\centering
\caption{\label{tab:multi_agent_results} Experiment 2 results}
\begin{tabular}{lc}
\hline
Metric & Value \\\hline
Agent 1 Bundle & Southwest, Moving Eastbound \\
Agent 2 Bundle & Northwest, Moving Southbound\\
Agent 3 Bundle & Northeast, Southeast\\
Completed tasks & 6/6 \\
Total mission duration & 20.89s \\
\hline
\end{tabular}
\end{minipage}

\end{table}

The environment includes only static obstacles. 
Table~\ref{tab:multi_agent_setup} describes the important parameters of the second experiment.  Table~\ref{tab:multi_agent_results} summarizes the results of the experiment, with the resulting trajectories depicted in Fig.~\ref{fig:multi_agent_environment_result}. Each task, including time-varying tasks, is accomplished by the team of agents.



\section{Conclusion} \label{sec:conclusion}

This paper presented a framework for coupling distributed CBBA task allocation with GCS motion planning in cluttered, dynamic environments. By integrating CBBA with ST-GCS, agents compute task bids from space-time feasible trajectories rather than from simplified distance-based estimates. This allows allocation decisions to reflect not only which tasks are nearby, but which task sequences are feasible and efficient.

The proposed framework also extends ST-GCS to multirotor planning in a 3D+time representation, where static obstacles, dynamic obstacles, moving tasks, and inter-agent conflicts can be modeled within a common space-time domain. The simulation results demonstrate that GCS-informed bidding can produce conflict-free trajectories that are consistent with the final motion-planning constraints, avoid dynamic obstacles, and complete both static and time-varying tasks. These results support the use of trajectory-aware bid generation as a mechanism for improving allocation quality in environments where geometric proximity alone is not a reliable indicator of execution cost or feasibility.



Future work will focus on improving both realism and scalability. On the coordination side, we will consider more realistic distributed communication models, including communication delays, packet loss, limited communication range, and asynchronous consensus updates. On the planning side, we will improve the computational performance of the ST-GCS backend through graph reuse, cost caching, warm starts, and more efficient convex-region generation. Additionally, the issues of agents colliding with each other and planning around other agent's bids will be addressed in future work. These improvements will be important for scaling the integrated CBBA+ST-GCS framework to larger teams, denser task sets, and more complex dynamic environments.

\section*{Acknowledgments}
This project was supported by NSF SBIR Phase 2 Award Number 2404858 and 4D Avionic Systems, LLC.  We thank Dr. Garth Thompson for his feedback and insightful suggestions.

During the preparation of this work, the authors used ChatGPT to improve the readability and wording of different sections of the paper. 
Additionally, ChatGPT was used to help prototype ideas and quickly iterate on code. After using ChatGPT, all writing and prototype code were extensively reviewed and edited/modified to ensure accuracy and coherence.  The authors take full responsibility for the content of the published article.

\bibliography{sample}

\end{document}